\documentclass[11pt]{article}
\usepackage{titletoc}
\usepackage[preprint]{acl}
\usepackage{times}
\usepackage{latexsym}
\usepackage[T1]{fontenc}
\usepackage[utf8]{inputenc}
\usepackage{microtype}
\usepackage{inconsolata}
\usepackage{booktabs} 
\usepackage{multirow}
\usepackage{graphicx}
\usepackage{amsmath}
\usepackage{amssymb}
\usepackage{capt-of}
\usepackage{pifont}
\usepackage{listings}
\usepackage{CJKutf8}
\usepackage{tabularx}
\usepackage{array}
\usepackage{xcolor}
\usepackage{subcaption}
\usepackage{makecell}
\definecolor{BrickRed}{RGB}{176,43,43}
\definecolor{ForestGreen}{RGB}{23,116,73}
\usepackage[most]{tcolorbox}
\tcbuselibrary{listings,breakable}
\definecolor{codegreen}{rgb}{0.0, 0.411, 0.243}
\definecolor{codered}{HTML}{DB4F59}
\usepackage{hyperref}
\definecolor{dartgreen}{HTML}{00693e}
\definecolor{refcolor}{HTML}{9F363A}
\definecolor{purple}{HTML}{999AC9}

\newtcblisting{promptbox}[1]{
  enhanced,
  breakable,
  listing only,
  listing engine=listings,
  colback=gray!3,
  colframe=gray!45,
  boxrule=0.4pt,
  arc=1pt,
  left=4pt,
  right=4pt,
  top=4pt,
  bottom=4pt,
  title={#1},
  fonttitle=\bfseries\footnotesize,
  halign=left,
  listing options={
    basicstyle=\ttfamily\scriptsize,
    breaklines=true,
    breakatwhitespace=true,
    breakindent=0pt,
    breakautoindent=false,
    columns=fullflexible,
    keepspaces=true,
    showstringspaces=false,
    resetmargins=true,
    xleftmargin=0pt,
    framexleftmargin=0pt,
    aboveskip=0pt,
    belowskip=0pt,
    tabsize=2
  }
}
\usepackage[table]{xcolor} \usepackage{booktabs} \usepackage{tabularx} \usepackage{array} 
\newcommand{\ours}{\textsc{\textsc{LoomSum}}}
\usepackage[dvipsnames]{xcolor}
\definecolor{quantblue}{RGB}{38,105,170}
\definecolor{linkgreen}{RGB}{25,130,85}
\definecolor{missingred}{RGB}{185,55,55}
\definecolor{relationred}{RGB}{180,35,35}
\definecolor{relationblue}{RGB}{0,80,150}

\newcommand{\wronglink}[1]{%
  \textcolor{relationred}{\textbf{#1}}%
}

\title{\textsc{LoomSum}: Weaving Quantitative and Narrative Evidence for Faithful Long Text–Table Summarization}
\author{
 \textbf{Meng Zhou}$^{1}$,
 \textbf{Wenhao You}$^{2}$,
 \textbf{Wei Yuan}$^{3}$,
 \textbf{Yueying Tian}$^{4}$
\\
 $^{1}$University of Toronto, 
 $^{2}$University of Waterloo,
  $^{3}$Independent Researcher,
  $^{4}$University of Sussex
 \\
  \texttt{simonzhou@cs.toronto.edu}, 
   \texttt{w22you@uwaterloo.ca}
}

\begin{document}
\maketitle
\begin{abstract}
Long documents often distribute important information across narrative passages and tables, making faithful summarization difficult. Existing methods may produce quantitative facts and analytical statements that are each supported by the source yet incorrectly linked, yielding summaries that appear numerically plausible but are analytically unfaithful. In this work, we propose \textsc{LoomSum}, a training-free framework that extracts source-grounded atomic claims, explicitly links table-derived facts to supporting narrative analyses, and plans discourse structure before generation using retrieved in-context demonstrations, without requiring any parameter updates. We further introduce Table-Grounded Faithfulness (TGF), a reference-free, claim-level metric that scores Numeric Grounding, Analysis Support, and Relation Consistency. On FINDSum, \textsc{LoomSum} obtains the highest TGF and grounded claim coverage among all compared methods; on USTT, it yields the highest numeric selection and grounded coverage, while an extractive baseline attains higher TGF by expressing fewer claims. In a human study, TGF components correlate in a positive direction with human judgments ($\rho$=0.31--0.63), and Relation Consistency correlates more strongly with human relation judgments than AlignScore and SummaC. Our results suggest that preserving relations between facts, not only grounding individual facts, is important for faithful long text--table summarization.

\end{abstract}

\section{Introduction} \label{sec:intro}

\begin{figure}[t]
\centering
\includegraphics[width=0.9\columnwidth]{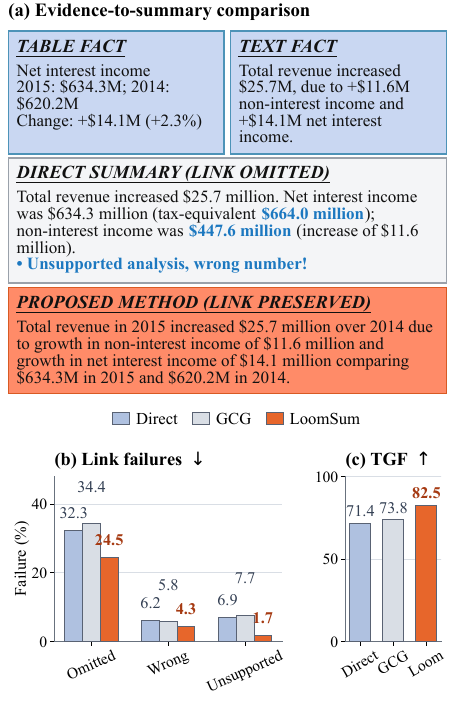}
\caption{Quantitative--narrative misalignment and its mitigation.
\textit{(a)} Direct summary omits a supported table--text link,
while \textsc{LoomSum} preserves it.
\textit{(b)} Link-error rates on a \textbf{\textsc{FINDSum-ROO} subset}.
\textit{(c)} TGF scores.}
\label{fig:motivation}
\end{figure}

Faithful summarization of long, heterogeneous documents remains difficult even for long-context large language models (LLMs). Larger context windows reduce truncation but do not ensure effective evidence use: models exhibit positional biases, unevenly use information across the input, and make more factual errors when relevant evidence appears in the middle of a document~\citep{liu2024lost,ravaut2024context,cao2024characterizing,wan2025positional}. Faithful long-document summarization therefore requires not only selecting salient information, but also integrating related evidence distributed across distant and heterogeneous source regions. This challenge is particularly pronounced in financial reports, where quantitative results are often presented in tables, while their causes, implications, and uncertainties are discussed in narrative passages elsewhere. Prior works such as FINDSum and USTT study this text--table setting~\citep{liu2022long,wang2023beyond}, but access to both modalities does not guarantee that they are correctly associated. Motivated by the numerical \emph{context mismatch} identified by~\citet{cao2024characterizing}, we isolate a cross-modal relational failure: \textit{a model may reproduce a source-supported quantity and a source-supported analytical statement while incorrectly linking them}. We call this failure \emph{quantitative--narrative misalignment}. Unlike conventional hallucination, both components may be individually grounded while the relation asserted between them is unsupported. Figure~\ref{fig:motivation} illustrates this error and previews our main finding: explicitly modeling cross-modal relations reduces omitted and incorrect links and improves Table-Grounded Faithfulness (TGF).

\noindent \textbf{Related Work. }Existing approaches only partially address this failure. Hierarchical, extract-then-abstract, and retrieval-based methods shorten long inputs by selecting salient sentences, table records, or relational facts before generation~\citep{chen2018fast,gehrmann2018bottom,zhu2021enhancing,cho2024rtsum,kirstein2025re}, and content planning decides what to say before realization~\citep{puduppully2021data}. For financial summarization, FINDSum selects textual segments and table tuples separately~\citep{liu2022long}, while~\citet{wang2023beyond} model sentence--table interactions. Hybrid table--text QA likewise shows that answers depend on chaining evidence across modalities~\citep{zhu2021tat,chen2021finqa}. However, these methods do not maintain the quantitative--narrative relation as an explicit, traceable unit throughout generation, and many require task-specific fine-tuning. Evaluation shares this gap. Entailment- and QA-based metrics score overall source--claim consistency~\citep{fabbri2022qafacteval,zha2023alignscore,laban2022summac}. Fine-grained analyses study error types and dependency-level entailment~\citep{goyal2020evaluating,pagnoni2021understanding,tang2023understanding}. Atomic-claim and LLM-judge methods verify decomposed claims~\citep{min2023factscore,song2024finesure,wei2024long,liu2023g}, and recent work adds attribution and coverage~\citep{wu2024less,chu2025tracsum,jeong2025agent,elaraby2026arc,samarinas2025beyond}. Numerical and table-aware metrics assess consistency with structured records or the selection of quantitative content~\citep{dhingra2019handling,wang2020towards,chen2020logical,liu2022long}. More recently, TabFaith~\citep{bukkapatnam2026tabfaith} introduces the STAF metric to verify claims against table cells, explicitly addressing numerical, relational, comparative, and temporal errors. While this work assesses structural faithfulness within tables, the challenge of linking quantities to their supported narrative interpretations is not fully addressed, motivating our explicit evaluation of quantitative--narrative relation consistency.

To address these limitations, we propose \textsc{LoomSum}, a training-free claim-to-discourse framework for long text--table summarization. It extracts source-grounded atomic claims from narrative text and tables, explicitly aligns quantities with evidence that explains, compares, qualifies, or contextualizes them, and consolidates the resulting claims into a discourse plan~\citep{puduppully2021data} for generation, from which an off-the-shelf LLM generates the final summary. We further introduce \textbf{Table-Grounded Faithfulness (TGF)}, a claim-level evaluation framework that separately assesses Numeric Grounding, Analysis Support, and Relation Consistency. This decomposition distinguishes errors in quantitative facts, narrative interpretations, and the relations between them. Experiments on FINDSum and USTT show that \textsc{LoomSum} improves analytical faithfulness while maintaining competitive summary quality and content coverage. A claim-level human
evaluation shows positive component-level associations with the corresponding judgments; notably, Relation Consistency agrees with human relation judgments more strongly than generic factuality metrics.

In summary, our contributions are as follows:
\begin{itemize}
    \item We identify and empirically characterize \textit{quantitative--narrative misalignment}, a relation-level error hidden when quantitative and narrative components are evaluated independently.
    \item We propose \textsc{LoomSum}, a training-free framework that constructs source-grounded atomic claims, explicitly aligns quantitative and narrative evidence, and organizes the claims through discourse planning for better summarization.
    \item We introduce and human-validate \textbf{TGF} score, a reference-summary-free, source-grounded metric that separately evaluates Numeric Grounding, Analysis Support, and Relation Consistency, capturing relation-level errors overlooked by existing factuality and numerical metrics.
    \item We conduct extensive experiments on FINDSum and USTT datasets, demonstrating improvements in analytical faithfulness and summary quality through automatic evaluation, ablation studies, and qualitative analyses.
\end{itemize}

\section{Method}

\subsection{Overview}
\begin{figure*}[t]
	\begin{center}
        \includegraphics[width=\textwidth]{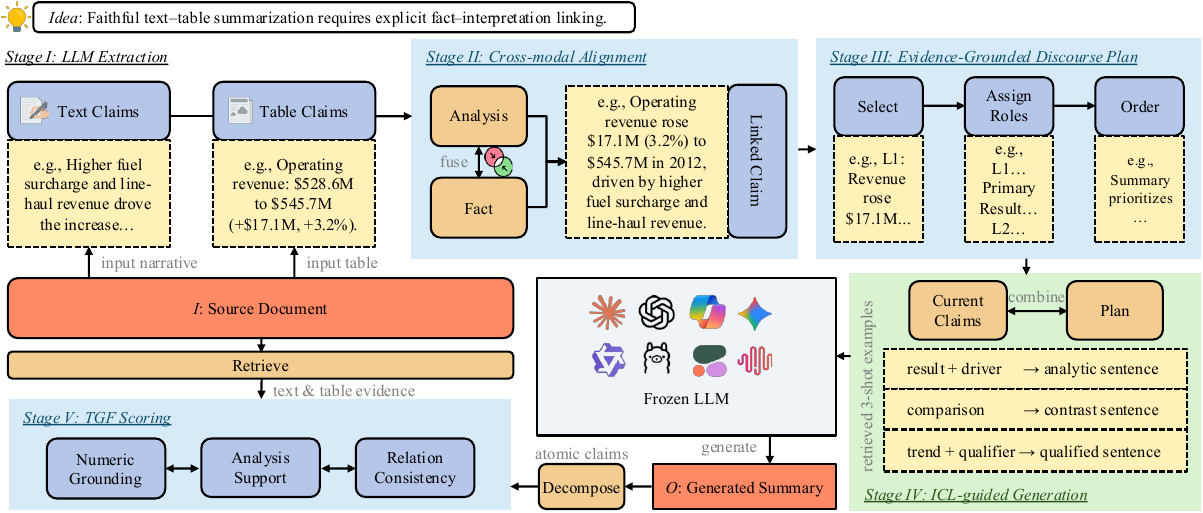}
	\end{center}
\caption{Overall view of our proposed method~\textsc{LoomSum}, which extracts source-grounded claims from text and tables, links quantitative facts to narrative interpretations, and organizes the aligned evidence into a discourse plan for summary generation.}
\label{loomsum_archi}
\end{figure*}

Given a source document $D$, the proposed method constructs source-grounded text and table claims, aligns quantitative facts with narrative interpretations, consolidates the resulting evidence into a discourse plan, and generates the final summary. The overall workflow is depicted in Figure~\ref{loomsum_archi}.

\subsection{Claim Construction and Cross-Modal Alignment}

\noindent \textbf{Claim Construction.} Inspired by claim-level decomposition used in FineSurE \citep{song2024finesure}, \ours converts the source into short, independently verifiable claims before generation: \refstepcounter{equation}\label{claim_con}\(\mathbf{C}_{\mathrm{text}},\mathbf{C}_{\mathrm{table}}=\operatorname{LLM}_{\mathrm{con}}\left(D^{\mathrm{text}},D^{\mathrm{table}}\right)\). For \textbf{text claims}, the LLM extracts atomic propositions while preserving entities, quantities, units, reporting periods, and stated relations. Each claim retains its supporting source span and document position, enabling subsequent alignment and provenance tracking without relying solely on the model-generated reformulation. For \textbf{table claims}, the model constructs atomic quantitative propositions from serialized table records that preserve the table title, row and column headers, value, unit, and reporting period. All claims retain source identifiers, which are propagated through alignment, consolidation, planning, and generation.

\noindent \textbf{Cross-modal Alignment.} Quantitative facts and their narrative interpretations may appear in distant regions of a long document. \ours therefore aligns table claims with text claims that explain, compare, qualify, attribute, or contextualize them:\refstepcounter{equation}\label{claim_align}
\(\mathbf{C}_{\mathrm{align}}=\operatorname{LLM}_{\mathrm{align}}(\mathbf{C}_{\mathrm{text}},\mathbf{C}_{\mathrm{table}})\). Each cross-modal analytical claim retains the quantitative component, the narrative component, their asserted relation, and the provenance of both. We also ensure that the alignment stage is constrained to link existing source-derived claims rather than introduce new factual content.

\subsection{Claim Consolidation and Planning}
\label{sec:consolidation_planning}
Long documents may repeat the same information across tables, narratives, and sections. To reduce redundancy, \ours conservatively consolidates claims: \refstepcounter{equation}\label{claim_conso}
\(\overline{\mathbf{C}}=\operatorname{LLM}_{\mathrm{conso}}(\mathbf{C}_{\mathrm{text}} \cup \mathbf{C}_{\mathrm{table}} \cup \mathbf{C}_{\mathrm{align}})\). The LLM is instructed to merge claims only when they express the same proposition and agree on their entity, metric, reporting period, etc. Claims with conflicting values, temporal scopes, or trends are retained separately. When several equivalent representations exist, a cross-modal analytic claim is preferred because it preserves both the quantitative fact and its narrative interpretation. Finally, following prior work on content planning \cite{puduppully2021data}, we organize the consolidated claims into an ordered discourse plan: \refstepcounter{equation}\label{plan}\(\mathbf{P}=\operatorname{LLM}_{\mathrm{plan}}(\overline{\mathbf{C}})\). The planner specifies which claims should be realized, their discourse roles, their supporting evidence, and the quantitative--narrative relations to preserve. Separating planning from surface realization promotes coherent information ordering
and reduces the risk of omitting or separating linked evidence during generation.

\subsection{Summary Generation}
The summary generator is guided by a small set of $K$ demonstrations, $\{X_i\}_{i=1}^{K}$, retrieved from the \textit{training split} through in-context learning. Each demonstration contains source-derived claims, their discourse organization, and the corresponding reference summary. These examples illustrate how evidence should be compressed, ordered, and combined while preserving quantitative--narrative relations. Given the discourse plan $\mathbf{P}$ and the retrieved
demonstrations, the generator produces the final summary: \refstepcounter{equation}\label{sum_gen}
\(\mathbf{Y}=\operatorname{LLM}_{\mathrm{gen}}(\mathbf{P}, \{X_i\})\). Each generated sentence is also accompanied by the identifiers of its supporting claims, providing lightweight sentence-level provenance and facilitating subsequent inspection.

\section{Table-Grounded Faithfulness Score}
\label{sec:tgf}

Existing metrics assess reference similarity, numerical content, or general claim--source consistency, but do not explicitly determine whether a quantitative fact is associated with the correct narrative interpretation. We therefore introduce \textbf{Table-Grounded Faithfulness (TGF)}, a \textit{reference-summary-free, source-grounded metric} with three dimensions: \emph{Numeric Grounding(NG)}, \emph{Analysis Support(AS)}, and \emph{Relation Consistency(RC)} to evaluate whether generated quantitative analyses are supported by the input document. Following the atomic-fact evaluation paradigm of FActScore \citep{min2023factscore}, TGF decomposes an LLM-generated summary $S$ into a set of independently verifiable claims, $\{c_i\}_{i=1}^{M}$, and retrieves an evidence packet $E_i$ from the corresponding source document $D$ for each claim. Claims with multiple independent propositions are split, except when a quantitative premise and its interpretation must remain together to evaluate their relation.

\noindent \textbf{LLM-based Scoring.} For each dimension $d\in\{NG,AS,RC\}$, an LLM judge (o3-mini\footnote{https://openai.com/index/openai-o3-mini/}) evaluates $c_i$ against $E_i$ using a five-point rubric. A score of $1$ denotes unsupported or contradicted content, whereas $5$ denotes full support; intermediate scores indicate increasing degrees of partial support. Because the judge does not provide calibrated probabilities over the five ratings, we repeat each judgment $L$ times. Let $\hat{s}_{i,\ell}^{(d)}\in\{1,\ldots,5\}$ denote the rating assigned to claim $c_i$ for dimension $d$ in the $\ell$-th judgment. We compute the normalized dimension score as $s_i^{(d)}
= \frac{1}{L}\sum_{\ell=1}^{L}\frac{\hat{s}_{i,\ell}^{(d)}-1}{4}.$
This maps each rating to a $[0,1]$ scale. We further set $N_i=s_i^{(NG)}$, $A_i=s_i^{(AS)}$, and $R_i=s_i^{(RC)}$.

\noindent \textbf{Evaluation dimensions.} \emph{Numeric Grounding} evaluates whether the quantitative content of $c_i$ is supported by $E_i$ together with its semantic context. The judge considers all attributes associated with it. Numerical surface forms are normalized only to resolve equivalent formats. A correct value associated with the wrong metric, entity, unit, or reporting period therefore does not receive full credit. \emph{Analysis Support} evaluates whether trends, comparisons,
interpretations, and explanations are licensed by the evidence $E_i$. Plausible but unstated interpretations receive lower scores than analyses explicitly supported by the source. \emph{Relation Consistency} applies to claims combining quantitative and narrative components and evaluates whether the source supports their asserted association. Relation Consistency is designed to capture this quantitative--narrative misalignment.

\noindent \textbf{Aggregation.} Each claim contributes once to TGF. We set $F_i=N_i$ for numeric-only
claims, $F_i=A_i$ for analysis-only claims, and $F_i=\min\{N_i,A_i,R_i\}$ for cross-modal analytic claims.
For a cross-modal analytical claim to be faithful, its quantitative fact, analytical explanation, and the relation connecting them must all be supported. We therefore adopt a non-compensatory aggregation in which the claim score is determined by its least-supported required component (the \texttt{min()} operator). This prevents strongly supported individual facts from compensating for an unsupported analytical link. Then, the summary-level score can be written as $\mathrm{TGF}(S)=\frac{1}{M}\sum_{i=1}^{M}F_i$, and the overall score across the test set is the document macro-average $\mathrm{TGF}_{\mathrm{test}} =\frac{1}{|D|}\sum_{i}\mathrm{TGF}(S_i)$.

\section{Experiments}

We organize our experiments around three research questions we aim to answer. \textbf{RQ1:} Does \ours improve summary quality, numeric content coverage, and table-grounded faithfulness over existing summarization methods?
\textbf{RQ2:} How do cross-modal alignment and planning contribute to the performance of \ours?
\textbf{RQ3:} Does TGF align with human judgments and capture relation-level errors that are overlooked by generic factuality metrics?

\subsection{Settings}
In this work, we evaluate \ours on two complementary table--text summarization benchmarks that differ in language and input scale. \textsc{FINDSum} is an English long-document, multi-table financial summarization dataset \citep{liu2022long}. It contains two subsets: \textbf{ROO}, which focuses on changes in revenue, expenses, and profitability, and \textbf{Liquidity}, which covers liquidity, cash flows, debt, and capital resources. We additionally evaluate on \textbf{USTT}, a Chinese financial table--text summarization dataset \citep{wang2023beyond}, in which each instance contains a table, an associated narrative passage, and a human-written summary grounded in both modalities. We use the official test split for all datasets. Detailed dataset statistics are provided in Appendix~\ref{app:ds_stats}.

\noindent \textbf{Baselines. }We compare \ours with methods that use different input modalities. For the text-only setting, \textsc{TextRank}~\citep{mihalcea2004textrank} is an unsupervised extractive baseline, while \textsc{BART}~\citep{lewis2020bart} and \textsc{BigBird-Pegasus}~\citep{zaheer2020big} are neural abstractive summarizers operating only on the narrative text. For the table-only setting, \textsc{BART} and \textsc{BigBird-Pegasus} verbalize selected table tuples, while \textsc{Table-RAG} retrieves relevant table rows before generation. For the joint text--table setting, \textsc{GCG} verbalizes table tuples and combines them with the narrative input for final generation, following~\citet{liu2022long}. \textsc{Direct-LC} provides the complete narrative and serialized tables directly to a long-context LLM, whereas \textsc{RAG-Sum} retrieves relevant narrative blocks and table rows using fixed summary-oriented queries. Unless otherwise stated, all LLM-based methods use GPT-5-nano~\citep{singh2025openai} as the base model and the same decoding configuration as \ours (\texttt{Reasoning Effort: Minimal; Max tokens: 8,000}). We additionally evaluate \textsc{LoomSum}'s sensitivity to both open-weight and proprietary base models. For ICL sample demonstrations, we retrieve exclusively from the training pool.

\noindent \textbf{Evaluation Metrics. }We evaluate summary quality using reference-based, numerical-content,
and source-grounded faithfulness metrics. For reference-based evaluation, we report ROUGE-1, ROUGE-2, and ROUGE-L scores \citep{lin2004rouge}, which measure unigram, bigram, and longest-common-subsequence overlap with the reference summary, respectively. We also report BERTScore \citep{zhang2019bertscore} to assess semantic similarity, following prior works \citep{liu2022long,wang2023beyond}. For numerical-content evaluation, we adopt Numerical Precision (NP), Numerical Coverage (NC), and their harmonic mean, Numerical Selection (NS), from \citet{liu2022long}. To evaluate source-grounded factual consistency, we include AlignScore \citep{zha2023alignscore} and our proposed TGF score, described in Section~\ref{sec:tgf}. Because TGF evaluates the faithfulness of expressed claims rather than omitted content, we additionally report Grounded Claim Coverage (G-Cov). We decompose the reference and generated summaries into atomic claims and measure the proportion of reference claims covered by at least one generated claim. G-Cov therefore complements the precision-oriented TGF by identifying methods that achieve high faithfulness through overly conservative generation. In the human validation study, we additionally include SummaC \citep{laban2022summac}. AlignScore and SummaC serve as generic factuality baselines, allowing us to assess whether TGF's dimension-specific design provides additional agreement with human judgments.

\section{Results}
\subsection{Results on FINDSum (RQ1)}

Table~\ref{tab:main_results} reports the main results on the whole ROO and Liquidity test sets of FINDSum. Due to space constraints, we report ROUGE-1 (R-1), ROUGE-2 (R-2), ROUGE-L (R-L), BERTScore (BS), Numeric Selection (NS), and Table-Grounded Faithfulness (TGF) in the main text; complete results and detailed analysis are provided in Appendix~\ref{app:full_quan_analysis}. Several patterns emerge from the results. First, LLM-based abstractive methods generally outperform unsupervised baselines across both reference-based and faithfulness-oriented metrics under all three input configurations. Text-only methods also generally outperform their table-only counterparts, suggesting that narrative passages provide a more directly usable representation of the content emphasized in the reference summaries. Among all methods, \ours performs best on most reported metrics across the two subsets. It nearly matches the best NS on ROO ($\Delta_{NS} = 0.05$ points(p)), while achieving the highest NS on Liquidity and the highest TGF on both subsets. The NS gains indicate improved retention of salient numerical content, while the TGF improvements show that quantities, analytical statements, and their asserted relations are more consistently supported by the source. These results support the central design of \ours: cross-modal alignment preserves quantitative--narrative relations, while discourse planning integrates the aligned evidence into a coherent summary.

\begin{figure}[t]
\centering
\includegraphics[width=0.9\linewidth]{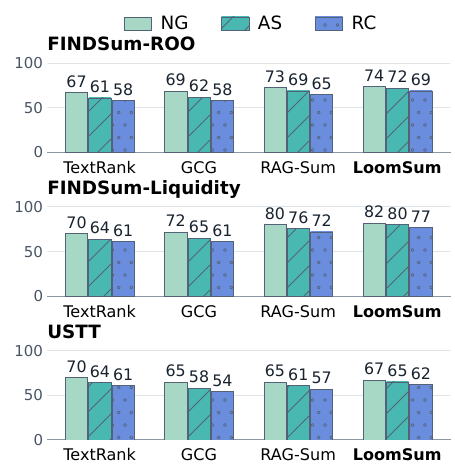}
\caption{Component-wise TGF scores across three datasets. Higher scores indicate better performance.}
\label{fig:tgf-dimensions}
\end{figure}

\noindent \textbf{Component-wise TGF analysis.} Figure~\ref{fig:tgf-dimensions} shows three component-level scores of TGF: Numeric Grounding (NG), Analysis Support (AS), and Relation Consistency (RC). On FINDSum-ROO, \ours achieves scores of 74 in NG, 72 in AS, and 69 in RC, respectively, outperforming the strongest baseline. The gains are larger on FINDSum-Liquidity, where \ours reaches 82 in NG, 80 in AS, and 77 in RC. The progressively larger gains from NG to RC indicate that \ours's primary advantage lies not only in reproducing grounded quantities, but also in preserving the analytical relations in which those quantities participate. Full results can be found in Appendix~\ref{app:full_quan_analysis}.

\begin{table*}[t]
\centering

\vspace{-1mm}
\begingroup
\footnotesize
\setlength{\tabcolsep}{2pt}
\renewcommand{\arraystretch}{0.96}

\begin{tabular*}{\textwidth}{
@{\extracolsep{\fill}}
ll
*{12}{c}
@{}
}
\toprule

\multirow{2}{*}{\textbf{Input}}
&
\multirow{2}{*}{\textbf{Method}}
&
\multicolumn{6}{c}{\textbf{FINDSum-Liquidity}}
&
\multicolumn{6}{c}{\textbf{FINDSum-ROO}}
\\

\cmidrule(lr){3-8}
\cmidrule(lr){9-14}

&
&
\textbf{R-1}
&
\textbf{R-2}
&
\textbf{R-L}
&
\textbf{BS}
&
\textbf{NS}
&
\textbf{TGF}
&
\textbf{R-1}
&
\textbf{R-2}
&
\textbf{R-L}
&
\textbf{BS}
&
\textbf{NS}
&
\textbf{TGF}
\\
\midrule

\multirow{3}{*}{\shortstack[l]{Text-\\only}}
& TextRank
& 33.06 & 6.98 & 13.54 & 74.33 & 20.05 & 65.86
& 35.04 & 7.34 & 14.94 & 76.81 & 24.80 & 63.19
\\

& BART
& 21.93 & 6.22 & 10.78 & 72.92 & 20.44 & 63.70
& 23.19 & 5.83 & 10.39 & 67.81 & 19.80 & 54.33
\\

& BigBird-Pegasus
& 25.84 & 7.86 & 12.59 & 73.80 & 21.40 & 64.28
& 24.10 & 7.01 & 13.49 & 68.24 & 22.80 & 64.37
\\
\midrule

\multirow{3}{*}{\shortstack[l]{Table-\\only}}
& BART
& 17.71 & 5.98 & 8.47 & 66.53 & 7.50 & 48.15
& 18.37 & 4.62 & 8.96 & 67.21 & 6.50 & 46.69
\\

& BigBird-Pegasus
& 23.16 & 6.11 & 11.50 & 68.50 & 13.34 & 50.92
& 19.25 & 6.90 & 12.91 & 69.70 & 12.47 & 52.22
\\

& Table-RAG$^\dagger$
& 30.05 & 5.88 & 12.70 & 65.59 & 14.68 & 48.92
& 31.80 & 7.32 & 14.91 & 69.72 & 12.27 & 51.93
\\
\midrule

\multirow{5}{*}{\shortstack[l]{Text +\\Table}}
& GCG$^\dagger$
& \textbf{33.96} & 7.79 & 14.14 & 76.54 & 24.59 & 66.71
& 29.25 & 8.89 & 16.58 & 80.53 & 23.75 & 64.47
\\

& Direct-LC$^\dagger$
& 32.01 & 7.49 & 13.74 & 75.38 & 25.11 & 69.29
& 30.19 & 8.83 & 16.52 & 80.55 & 24.10 & 68.21
\\

& RAG-Sum$^\dagger$
& 33.51 & 7.72 & 14.30 & 76.31 & 23.48 & 76.34
& 27.82 & 8.04 & 15.80 & 80.70 & \textbf{27.08} & 69.47
\\

& \textbf{\ours (Ours)}$^\dagger$
& 32.74 & \textbf{9.51} & \textbf{16.35}
& \textbf{76.78} & \textbf{25.98} & \textbf{79.43}
& \textbf{35.45}
& \textbf{10.77}
& \textbf{18.55}
& \textbf{81.34}
& 27.03
& \textbf{71.38}
\\

\bottomrule
\end{tabular*}

\endgroup
\caption{
Main results on the FINDSum-Liquidity and FINDSum-ROO test sets. Higher values are better. $\dagger$ denotes methods using GPT-5-nano as the base model. Best results are shown in \textbf{bold}
}
\label{tab:main_results}
\end{table*}

\subsection{Results on USTT (RQ1)}

Table~\ref{tab:ustt_results} reports the results on USTT. We observe similar findings to those on FINDSum. First, Table-only methods perform substantially worse than text-based and multimodal counterparts, suggesting that tabular records alone provide insufficient context for recovering the narrative structure and analytical content emphasized in the reference summaries. \ours achieves the highest R-1, R-L, and NS scores, indicating stronger overall content coverage and substantially better retention of reference-relevant numerical information.

An interesting result is that TextRank obtains the highest TGF score despite considerably lower ROUGE and NS performance. This reflects the distinction between faithfulness and coverage: an extractive method can achieve high claim-level faithfulness by selecting a small set of directly supported sentences while omitting other salient information. The component scores in the bottom of Figure~\ref{fig:tgf-dimensions} clarify this result. TextRank achieves the highest NG score, whereas \ours obtains the highest AS and RC scores. Thus, TextRank is effective at preserving isolated source quantities, while \ours provides stronger analytical support and more reliable quantitative--narrative integration.

Across all three evaluation sets, RC is generally the most challenging dimension. Nevertheless, \ours consistently achieves the highest RC score, providing direct evidence that explicit cross-modal alignment mitigates the quantitative--narrative misalignment identified in Section~\ref{sec:intro}. More detailed analysis can be found in Appendix~\ref{app:full_quan_analysis}.

\begin{table*}[t]
\centering

\begin{minipage}[t]{0.485\textwidth}
\vspace{0pt}
\centering
\scriptsize
\setlength{\tabcolsep}{1.8pt}
\renewcommand{\arraystretch}{0.94}

\resizebox{\linewidth}{!}{
\begin{tabular}{@{}llcccccc@{}}
\toprule
\textbf{Input}
& \textbf{Method}
& \textbf{R-1}
& \textbf{R-2}
& \textbf{R-L}
& \textbf{BS}
& \textbf{NS}
& \textbf{TGF} \\
\midrule

\multirow{3}{*}{\shortstack[l]{Text-\\only}}
& TextRank
& 24.77 & \textbf{10.20} & 13.83 & 60.24 & 21.44 & \textbf{66.41}\\
& BART
& 22.84 & 8.90 & 14.25 & 59.80 & 19.12 & 54.26 \\
& BigBird-Pegasus
& 24.18 & 7.69 & 14.17 & 60.12 & 20.35 & 56.91 \\
\midrule

\multirow{3}{*}{\shortstack[l]{Table-\\only}}
& BART
& 20.46 & 5.21 & 13.72 & 60.01 & 13.32 & 52.53 \\
& BigBird-Pegasus
& 20.71 & 5.54 & 12.01 & 59.25 & 13.98 & 55.79 \\
& Table-RAG$\dagger$
& 19.26 & 5.66 & 10.77 & 61.50 & 12.18 & 56.13 \\
\midrule

\multirow{4}{*}{\shortstack[l]{Text +\\Table}}
& GCG$\dagger$
& 24.83 & 8.93 & 12.92 & 63.84  & 18.76 & 59.71  \\
& Direct-LC$\dagger$
& 25.14 & 9.37 & 13.26 & 63.71  & 18.63  & 58.51  \\
& RAG-Sum$\dagger$
& 24.56 & 9.91 & 12.93 & \textbf{64.47} & 15.89 & 60.64  \\
& \textbf{\ours (Ours)}$\dagger$
& \textbf{26.10}
& 9.18
& \textbf{15.16}
& 64.16
& \textbf{27.36}
& 63.71 \\
\bottomrule
\end{tabular}
}
\captionof{table}{
Main results on the USTT test set. Higher values are better. $\dagger$ denotes methods using GPT-5-nano as the base model. Best results are shown in \textbf{bold}.
}
\label{tab:ustt_results}
\end{minipage}\hspace{0.02\textwidth}%
\begin{minipage}[t]{0.485\textwidth}
\vspace{0pt}
\centering

\scriptsize
\setlength{\tabcolsep}{2.0pt}
\renewcommand{\arraystretch}{0.82}

\resizebox{\linewidth}{!}{
\begin{tabular}{@{}lcccccc@{}}
\toprule
\textbf{Variant}
& \textbf{Align}
& \textbf{Plan}
& \textbf{ICL}
& \textbf{R-L}
& \textbf{NS}
& \textbf{TGF} \\
\midrule

\multicolumn{7}{@{}l}{
  \rule{0pt}{1.75ex}\textbf{FINDSum--ROO}
} \\
\cmidrule(lr){1-7}

Flat Evidence
& \ding{55} & \ding{55} & \ding{55}
& 13.47 & 21.46 & 55.68 \\

+ cross-modal Alignment
& \checkmark & \ding{55} & \ding{55}
& 15.89 & 22.30 & 62.93 \\

+ Planning
& \checkmark & \checkmark & \ding{55}
& 16.04 & 25.71 & 67.34 \\

\textbf{\ours (Ours)}
& \checkmark & \checkmark & \checkmark
& \textbf{18.55}
& \textbf{27.03}
& \textbf{71.38} \\

\midrule

\multicolumn{7}{@{}l}{
  \rule{0pt}{1.75ex}\textbf{FINDSum--Liquidity}
} \\
\cmidrule(lr){1-7}

Flat Evidence
& \ding{55} & \ding{55} & \ding{55}
& 13.12 & 20.21 & 58.94 \\

+ cross-modal Alignment
& \checkmark & \ding{55} & \ding{55}
& 14.47 & 21.47 & 65.69 \\

+ Planning
& \checkmark & \checkmark & \ding{55}
& \textbf{16.44} & 24.16 & 74.48 \\

\textbf{\ours (Ours)}
& \checkmark & \checkmark & \checkmark
& 16.35
& \textbf{25.98}
& \textbf{79.43} \\

\bottomrule
\end{tabular}
}
\captionof{table}{
Component ablation on FINDSum datasets. Best results are shown in \textbf{bold}.
}
\label{tab:component_ablation}
\end{minipage}

\end{table*}

\subsection{Results on Human Evaluation (RQ3)}

We conduct a human evaluation to examine whether the three components of TGF--NG, AS, and RC--agree with human judgments of the corresponding properties. We evaluate each dimension independently to align with the multidimensional design of TGF. More details about human evaluations are provided in Appendix~\ref{app:human_eval}.

\noindent \textbf{Annotation setup.} We randomly select 150 samples (test instances) and their generated summaries from the FINDSum test sets. After decomposing the generated summaries into atomic claims, we construct separate candidate pools for NG, AS, and RC. We then sample dimension-specific claim sets while balancing the dataset subset and generation method, yielding 167 NG, 171 AS, and 139 RC claims. The study is conducted by three annotators. NG is evaluated for claims containing quantitative information, AS for claims containing analytical statements, and RC for claims that associate a quantitative fact with a narrative interpretation. Each annotator independently assesses the applicable dimensions using the corresponding source evidence, including relevant table records and narrative passages. Annotators are blinded to the generation methods and all automatic metric scores. Cases of maximal disagreement, defined as when a claim receives final ratings of 1 and 5, are resolved through discussion among all annotators to produce a consensus score.

\begin{figure}[t]
\centering
\includegraphics[width=\columnwidth]{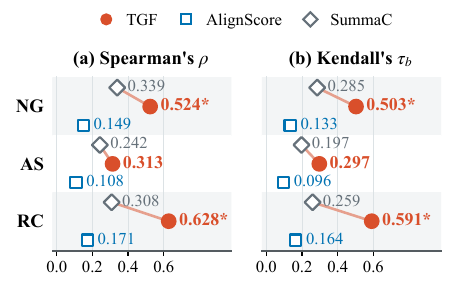}
\caption{Correlation of TGF and generic factuality metrics with human judgments across the three annotated dimensions.}
\label{fig:tgf-dimensions-stats}
\end{figure}

\noindent \textbf{How well does TGF align with human judgments?} We assess the validity of each TGF component by computing Spearman's $\rho$ and Kendall's $\tau_b$ between the LLM-derived scores defined in Section~\ref{sec:tgf} and the corresponding human scores. As shown in Figure~\ref{fig:tgf-dimensions-stats}, TGF-NG exhibits a moderate positive correlation with human Numeric Grounding judgments ($\rho=0.524$, $\tau_b=0.503$, $p<0.05$). TGF-RC achieves the strongest correlation ($\rho=0.628$, $\tau_b=0.591$, $p<0.05$), indicating that it reliably reflects human assessments of whether quantitative facts are associated with the correct narrative interpretations. TGF-AS shows a positive correlation with human Analysis Support judgments ($\rho = 0.313$, $\tau_b = 0.297$), with the lowest observed correlations among the three dimensions. Overall, the results provide human validation for the NG and RC components, with particularly strong evidence for RC. The weaker AS correlation suggests that determining whether an interpretation is sufficiently supported remains more subjective and challenging for both human and automatic evaluation.

\noindent \textbf{Does TGF capture human judgments beyond generic factuality metrics?} We additionally compare TGF with AlignScore and SummaC, two generic factuality metrics that each produce a single overall consistency score. For a fair comparison, we correlate each metric with Human NG, AS, and RC on the same annotated samples. Figure~\ref{fig:tgf-dimensions-stats} shows that TGF consistently achieves stronger agreement with human judgments than the generic baselines across both Spearman's $\rho$ and Kendall's $\tau_b$, with \textbf{*} representing $p<0.05$. The advantage is especially pronounced for Relation Consistency, where TGF substantially outperforms AlignScore and SummaC. This result supports our central claim that generic source--claim consistency metrics are not sufficient to detect \emph{quantitative--narrative relation errors}. TGF-RC explicitly targets this failure mode, which explains its markedly stronger correspondence with human RC judgments. We also observe gains for NG and AS, suggesting that the component-specific design of TGF provides more fine-grained alignment with human assessment than a single undifferentiated factuality score.

\subsection{Ablations (RQ2)}
Table~\ref{tab:component_ablation} adds components cumulatively to \emph{Flat Evidence} (independently extracted text and table claims, with no alignment, planning, or demonstrations). Each component improves TGF and NS on both subsets (ROO/Liquidity TGF: alignment {\color{orange}+7.3p}/{\color{orange}+6.8p}, planning {\color{orange}+4.4p}/{\color{orange}+8.8p}, demonstrations {\color{orange}+4.0p}/{\color{orange}+5.0p}). This suggests that linking table facts to narrative evidence and organizing the linked claims before generation both help preserve quantitative--narrative relations. R-L is essentially unchanged at the last step on Liquidity (16.44 vs.\ 16.35). \textbf{Do demonstrations help?} Since demonstrations contribute substantially, we give Direct-LC and RAG-Sum the same retrieved demonstrations. Demonstrations help the baselines only slightly (TGF {\color{orange}+0.9p} for Direct-LC to {\color{orange}+1.6p} for RAG-Sum), compared with {\color{orange}+4.0p}/{\color{orange}+5.0p} for \ours on ROO/Liquidity. \ours keeps the best R-L and TGF on both subsets, but the margin in TGF over the strongest matched baseline (RAG-Sum+ICL) is marginal ({\color{orange}+1.0p} on ROO, {\color{orange}+2.0p} on Liquidity) over the test set. NS is on par (ROO: 27.03 vs.\ 27.11 for RAG-Sum+ICL; Liquidity: 25.98 vs.\ 25.88 for Direct-LC+ICL). \textbf{Sensitivity Analysis: }We evaluate \ours with two proprietary and two open-weight base models. Across the evaluated backbones, the selected metrics remain relatively stable; this suggests that \ours is relatively robust to the choice of base model. Full results are provided in Appendix~\ref{app:base_model_analysis}.

\subsection{Case Study}
Finally, to complement the main results, we examine an example from the FINDSum-ROO test set. Table~\ref{tab:ajx-link-case} illustrates a quantitative--narrative
relation in which a \$10.8 million provision benefit increases net
interest income from \$51.4 million to \$62.2 million and is attributed
to the mortgage-loan portfolio and beneficial-interest investments.
While the baselines omit or weaken parts of this relation,
\textsc{LoomSum} preserves both the driver--outcome link and its
portfolio-level attribution. More examples are provided in
Appendix~\ref{sec:app_case_study}.

\begin{table}[t]
\centering
\scriptsize
\setlength{\tabcolsep}{2.5pt}
\renewcommand{\arraystretch}{1.03}

\newcommand{\linkmiss}{%
  \textcolor{red!65!black}{\(\boldsymbol{\times}\)}}
\newcommand{\linkkeep}{%
  \textcolor{green!45!black}{\(\boldsymbol{\checkmark}\)}}

\begin{tabularx}{\columnwidth}{
  @{}
  >{\raggedright\arraybackslash}p{0.18\columnwidth}
  >{\raggedright\arraybackslash}X
  @{}
}
\toprule

\multicolumn{2}{
  @{}>{\raggedright\arraybackslash}p{\columnwidth}@{}
}{
\textbf{Source evidence.}
\textbf{Table:} Net interest income: 51.4; provision benefit:
+10.8; post-provision net interest income: 62.2.
\textbf{Text:} ``... credit loss provisions of \$10.8 million
on our mortgage loan portfolio and investments in beneficial
interests during the year ended December 31, 2020.''
} \\

\midrule

Direct-LC \linkmiss
&
``... initial credit loss provisions of \$5.1 million in Q1 2020, the company recorded a net
recovery of \$10.8 million in credit loss provisions for the full year 2020 due to better-than-expected loan
performance. ...''
\\

GCG \linkmiss
&
``... credit loss provisions of \$10.8 million on mortgage loans
and beneficial interests for the year ended December 31, 2020;
net \$0.9 million of unrealized losses ...''
\\

RAG-Sum \linkmiss
&
``... macroeconomic conditions and expected lifetime credit
losses on our mortgage loan and beneficial interest portfolios
are subject to meaningful uncertainty ...''
\\

\textbf{\ours} \linkkeep
&
``... net interest income after provision for credit
benefit/(losses) for 2020 was \textbf{\$62.2 million}, compared
with the original \$51.4 million, reflecting the
\textbf{\$10.8 million} provision benefit recorded on our
mortgage loan portfolio and investments in beneficial
interests ...''
\\

\midrule

Reference
&
``... our net interest income after the recovery of provision
for credit losses increased to \textbf{\$62.2 million} [...]
primarily as a result of \textbf{\$10.8 million} in net recovery
of provision for credit losses ...''
\\

\bottomrule
\end{tabularx}

\caption{
Summarization comparison on a FINDSum-ROO example. \linkmiss{} denotes an omitted or wrong quantitative--narrative link, whereas \linkkeep{} denotes a preserved correct link.
}
\label{tab:ajx-link-case}
\end{table}

\section{Conclusion}
In this work, we identify \emph{quantitative--narrative misalignment}, where individually grounded facts and interpretations are combined into unsupported relations. We introduce \textsc{LoomSum}, which links table-derived quantities with narrative evidence before discourse planning and generation. Across FINDSum and USTT, \textsc{LoomSum} improves analytical faithfulness while maintaining summary quality. We also introduce TGF, whose Relation Consistency component aligns more strongly with human judgments than generic factuality metrics. These results suggest that preserving relations between facts, not only grounding individual facts, is important for faithful text--table summarization. 

\section*{Limitations}

This work provides an initial investigation of quantitative--narrative misalignment in long text--table
summarization. \ours is intentionally designed as a lightweight, training-free framework to isolate the value of explicitly linking quantitative facts with their narrative interpretations. It should therefore be viewed as one possible instantiation of relation-aware summarization rather than an
exhaustive solution to the problem. Its modular pipeline may also propagate errors across evidence extraction, cross-modal alignment, discourse planning, and generation. Future work could explore jointly optimized or learned relation models, richer evidence graphs, and generation objectives that directly preserve cross-modal relations.

Our empirical evaluation is conducted on FINDSum and USTT, both of which contain financial text--table inputs. Although the underlying failure mode may arise in other domains that combine structured records with narrative explanations, such as scientific, medical, and policy documents, its prevalence and characteristics outside financial reporting remain to be established. Extending the evaluation to broader domains, languages, table structures, and document lengths is therefore an important direction for future work.

Finally, TGF relies on automated claim extraction and LLM-based judgments, which may introduce model-dependent errors and sensitivity to evidence presentation. Our human study provides positive validation for Numeric Grounding and Relation Consistency, but is limited in scale, and the weaker association observed for
Analysis Support suggests that analytical interpretations remain more difficult to evaluate consistently. In addition, TGF is precision-oriented and evaluates the faithfulness of expressed claims rather than the omission of salient content. Future work could combine relation-aware faithfulness with source-level coverage evaluation and validate the metric using larger and more diverse human annotations.

\section*{Ethical Considerations}

Although we use datasets related to financial summaries that may inform consequential analytical or investment decisions, \ours is intended as a research and decision-support tool, not as a substitute for expert review or the sole input to automated decision-making. Errors in evidence extraction, cross-modal alignment, or generation may still produce plausible but misleading explanations. While our experiments use public benchmark data, applying the framework to non-public documents through proprietary LLM APIs may raise confidentiality and data-governance concerns. Reliance on a proprietary model for TGF also introduces limited transparency, model-version dependence, and reproducibility challenges. Practical deployments should retain source provenance, require human verification for consequential uses, document model versions and configurations, and use secure or locally hosted models where appropriate. Although no model training is required, repeated LLM inference still incurs computational and environmental costs.

\bibliography{custom}

\newpage

\appendix

\startcontents[appendices]
\section*{Contents of Appendix}
\printcontents[appendices]{}{1}{\normalsize}


\section{Datasets Statistics} \label{app:ds_stats}
Table~\ref{tab:app_dataset_statistics} summarizes the statistics of the \textbf{test sets} used in our experiments. We report the number of instances, the average source length, and the average reference summary length for each dataset. All lengths are measured in tokens using the GPT o200k\_base tokenizer. For FINDSum, the source length includes both the narrative content and the associated tabular information.

The two FINDSum subsets represent substantially longer inputs than USTT. FINDSum-Liquidity contains 4,204 test instances, with an average source length of 34,630 tokens and an average reference summary length of 538 tokens. FINDSum-ROO contains a larger test set of 6,201 instances and slightly longer source documents, averaging 36,447 tokens, while its reference summaries are somewhat shorter, with an average length of 444 tokens. These statistics highlight the long-context nature of FINDSum, where a summarization method must
identify and integrate salient evidence distributed across extensive narrative passages and multiple tables.

In contrast, USTT contains 1,390 test instances, with an average source length of 1,680 tokens and an average summary length of 85 tokens. Although USTT is considerably shorter than FINDSum, it provides a complementary table--text summarization setting with a different language and input scale. Evaluating on both datasets therefore allows us to examine whether the proposed method is effective not only for long financial reports, but also for shorter table--text inputs.

\begin{table}[h]
\centering
\footnotesize
\setlength{\tabcolsep}{3pt}
\renewcommand{\arraystretch}{1.0}
\begin{tabular}{@{}lrrr@{}}
\toprule
\textbf{Dataset}
& \textbf{\#Inst.}
& \textbf{Avg.\ Doc.}
& \textbf{Avg.\ Sum.} \\
\midrule
FINDSum-Liquidity     & 4204 & 34630 & 538 \\
FINDSum-ROO & 6201 & 36447 & 444 \\
USTT    &  1390 & 1680 & 85 \\
\bottomrule
\end{tabular}
\caption{Dataset statistics. Lengths are measured in tokens.}
\label{tab:app_dataset_statistics}
\end{table}

\section{More on Quantitative Results} \label{app:full_quan_analysis}

\begin{figure*}[t]
\centering
\includegraphics[width=\textwidth]{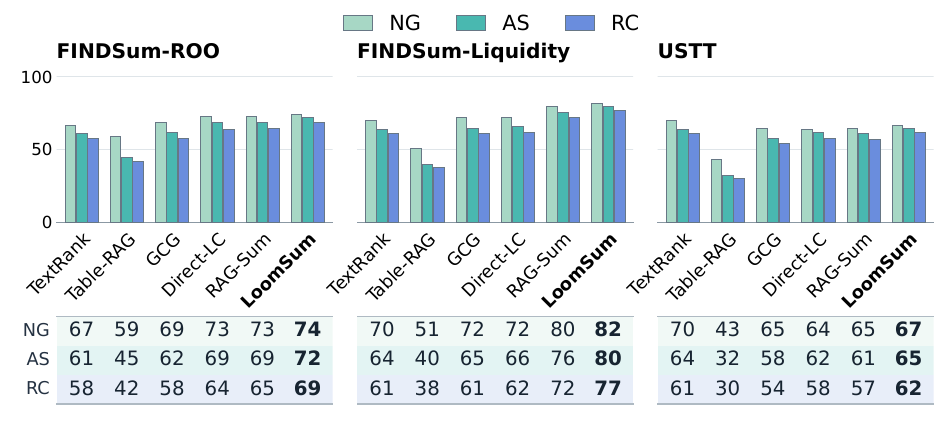}
\caption{Component-wise TGF scores across three
datasets for the selected methods. Higher scores indicate better performance. Values are scaled in the range of 0 to 100.}
\label{fig:tgf-dimensions-full}
\end{figure*}

Tables~\ref{tab:app_main_results_full_roo} and~\ref{tab:app_main_results_full_liq} report the complete
quantitative results on the FINDSum-ROO and FINDSum-Liquidity test sets, respectively. Since both datasets are large, we only run once, following the official setup in \citet{liu2022long} and \citet{wang2023beyond}. In addition to the metrics reported in the main text, we include Numerical Precision (NP), Numerical Coverage (NC), AlignScore, SummaC, and Grounded Claim Coverage (G-Cov). The full component-wise TGF scores for all methods are shown in Figure~\ref{fig:tgf-dimensions-full}.

\noindent\textbf{Grounded Claim Coverage (G-Cov).} G-Cov is a reference-based grounded coverage metric that
complements the precision-oriented TGF. TGF evaluates the faithfulness of the claims expressed by a method and averages over the generated claim set. \textit{Consequently, a short or conservative summary may obtain a high TGF by expressing only a small number of well-supported claims}. G-Cov instead uses claims from the reference summary and measures how much salient reference content is both recovered by the generated summary and grounded in the source.

For each document $D$, we decompose the reference and generated summaries into atomic claim sets
$\mathcal{C}_{\mathrm{ref}}(D)$ and $\mathcal{C}_{\mathrm{gen}}(D)$ as done in TGF. For each reference claim $r_i\in\mathcal{C}_{\mathrm{ref}}(D)$, we select its best-aligned generated claim:
\begin{equation}
g_i^{*}
=
\arg\max_{g_j\in\mathcal{C}_{\mathrm{gen}}(D)}
A(r_i,g_j),
\label{eq:gcov_matching}
\end{equation}
where $A(r_i,g_j)\in\{1,\ldots,5\}$ is an LLM-as-Judge-based semantic alignment rating. Alignment evaluates whether the generated claim preserves the content of the reference claim, including its entities, metrics, values, periods, directions, and asserted relations. The normalized alignment credit is
\begin{equation}
a_i
=
\frac{A(r_i,g_i^{*})-1}{4}.
\label{eq:gcov_alignment}
\end{equation}

The matched generated claim is additionally evaluated for Numeric Grounding (NG), Analysis Support (AS), and Relation Consistency (RC) against the source evidence. Each applicable rating is normalized to $[0,1]$:
\begin{equation}
\widetilde{S}_i
=
\frac{S_i-1}{4},
\qquad
S\in\{\mathrm{NG},\mathrm{AS},\mathrm{RC}\}.
\label{eq:gcov_normalization}
\end{equation}
According to the type of the reference claim, its faithfulness credit is defined as
\begin{equation}
f_i =
\begin{cases}
\widetilde{\mathrm{NG}}_i,
& r_i \text{ is numeric},\\[2pt]
\widetilde{\mathrm{AS}}_i,
& r_i \text{ is analytical},\\[2pt]
\min\{
\widetilde{\mathrm{NG}}_i,
\widetilde{\mathrm{AS}}_i,
\widetilde{\mathrm{RC}}_i
\},
& r_i \text{ is cross-modal}.
\end{cases}
\label{eq:gcov_faithfulness}
\end{equation}
Then, the grounded coverage credit assigned to $r_i$ is
\begin{equation}
c_i=a_i f_i,
\label{eq:gcov_credit}
\end{equation}
and the document-level G-Cov score is
\begin{equation}
\operatorname{G\text{-}Cov}(D)
=
\frac{1}{|\mathcal{C}_{\mathrm{ref}}(D)|}
\sum_{r_i\in\mathcal{C}_{\mathrm{ref}}(D)}
c_i.
\label{eq:gcov_document}
\end{equation}
The final dataset score is obtained by averaging over documents.

Alignment and faithfulness serve distinct roles in this definition. A generated claim may resemble a reference claim but remain unsupported by the source; conversely, it may be source-grounded but express content unrelated to the reference claim. A reference claim receives high G-Cov credit only when both semantic coverage and source grounding are strong. G-Cov should therefore be interpreted as grounded recall of the salient content represented by the reference summary, rather than exhaustive coverage of all information in the source document.

\begin{table*}[t]
\centering
\scriptsize
\setlength{\tabcolsep}{2.2pt}
\renewcommand{\arraystretch}{0.96}

\resizebox{\textwidth}{!}{
\begin{tabular}{@{}llccccccccccc@{}}
\toprule
\textbf{Input}
& \textbf{Method}
& \textbf{R-1}
& \textbf{R-2}
& \textbf{R-L}
& \textbf{BS}
& \textbf{NP}
& \textbf{NC}
& \textbf{NS}
& \textbf{AlignScore}
& \textbf{SummaC}
& \textbf{G-Cov}
& \textbf{TGF} \\
\midrule

\multirow{3}{*}{\shortstack[l]{Text-\\only}}
& TextRank
& 35.04 & 7.34 & 14.94 & 76.81
& 24.65 & 40.90 & 24.80 & 90.68 & 54.96 & 12.26 & 63.19 \\
& BART
& 23.19 & 5.83 & 10.39 & 67.81
& 22.41 & 21.56 & 19.80 & 94.63 & 81.82 & 10.73 & 54.33 \\
& BigBird-Pegasus
& 24.10 & 7.01 & 13.49 & 68.24
& 18.17 & 31.29 & 22.80 & 86.78 & 64.09 & 10.54 & 64.37 \\
\midrule

\multirow{3}{*}{\shortstack[l]{Table-\\only}}
& BART
& 18.37 & 4.62 & 8.96 & 67.21
& 13.24 & 6.70 & 6.50 & 56.96 & 34.38 & 10.51 & 46.69 \\
& BigBird-Pegasus
& 19.25 & 6.90 & 12.91 & 69.70
& 9.98 & 20.43 & 12.47 & 50.45 & 52.28 & 9.02 & 52.22 \\
& Table-RAG$^\dagger$
& 31.80 & 7.32 & 14.91 & 69.72
& 10.46 & 18.86 & 12.27 & 22.67 & 44.84 & 11.71 & 51.93 \\
\midrule

\multirow{4}{*}{\shortstack[l]{Text +\\Table}}
& GCG$^\dagger$
& 29.25 & 8.89 & 16.58 & 80.53
& 22.64 & 30.49 & 23.75 & \textbf{95.31} & 62.07 & 13.57 & 64.47 \\
& Direct-LC$^\dagger$
& 30.19 & 8.83 & 16.52 & 80.55
& 21.27 & 29.02 & 24.10 & 92.66 & 85.92 & 14.72 & 68.21 \\
& RAG-Sum$^\dagger$
& 27.82 & 8.04 & 15.80 & 80.70
& \textbf{25.57} & 42.54 & \textbf{27.08} & 88.21 & 85.93 & 13.10 & 69.47 \\
& \textbf{\ours (Ours)}$^\dagger$
& \textbf{35.45}
& \textbf{10.77}
& \textbf{18.55}
& \textbf{81.34}
& 22.53 & \textbf{54.17}
& 27.03
& 90.28 & \textbf{86.46} & \textbf{17.75}
& \textbf{71.38} \\
\bottomrule
\end{tabular}
}
\caption{
Full results on the FINDSum-ROO test set. Higher values are better.
$\dagger$ denotes methods using GPT-5-nano as the backbone.
Best results are shown in \textbf{bold}. All metrics are reported on a 0--100 scale.
}
\label{tab:app_main_results_full_roo}
\end{table*}

\begin{table*}[t]
\centering
\scriptsize
\setlength{\tabcolsep}{2.2pt}
\renewcommand{\arraystretch}{0.96}

\resizebox{\textwidth}{!}{
\begin{tabular}{@{}llccccccccccc@{}}
\toprule
\textbf{Input}
& \textbf{Method}
& \textbf{R-1}
& \textbf{R-2}
& \textbf{R-L}
& \textbf{BS}
& \textbf{NP}
& \textbf{NC}
& \textbf{NS}
& \textbf{AlignScore}
& \textbf{SummaC}
& \textbf{G-Cov}
& \textbf{TGF} \\
\midrule

\multirow{3}{*}{\shortstack[l]{Text-\\only}}
& TextRank
& 33.06
& 6.98
& 13.54
& 74.33
& 17.18
& 42.37
& 20.05
& 80.53
& 62.07
& 12.06
& 65.86
\\

& BART
& 21.93
& 6.22
& 10.78
& 72.92
& 21.89
& 24.20
& 20.44
& \textbf{85.74}
& \textbf{72.49}
& 10.56
& 63.70
\\

& BigBird-Pegasus
& 25.84
& 7.86
& 12.59
& 73.80
& \textbf{22.33}
& 23.97
& 21.40
& 83.17
& 60.20
& 10.24
& 64.28
\\
\midrule

\multirow{3}{*}{\shortstack[l]{Table-\\only}}
& BART
& 17.71
& 5.98
& 8.47
& 66.53
& 11.68
& 8.53
& 7.50
& 56.25
& 33.24
& 9.20
& 48.15
\\

& BigBird-Pegasus
& 23.16
& 6.11
& 11.50
& 68.50
& 9.43
& 31.25
& 13.34
& 59.53
& 54.68
& 10.05
& 50.92
\\

& Table-RAG$^\dagger$
& 30.05
& 5.88
& 12.70
& 65.59
& 10.70
& 33.95
& 14.68
& 28.69
& 44.56
& 11.20
& 48.92
\\
\midrule

\multirow{4}{*}{\shortstack[l]{Text +\\Table}}
& GCG$^\dagger$
& \textbf{33.96}
& 7.79
& 14.14
& 76.54
& 16.87
& 51.45
& 24.59
& 78.19
& 63.05
& 16.25
& 66.71
\\

& Direct-LC$^\dagger$
& 32.01
& 7.49
& 13.74
& 75.38
& 17.29
& 54.78
& 25.11
& 79.22
& 64.29
& 17.75
& 69.29
\\

& RAG-Sum$^\dagger$
& 33.51
& 7.72
& 14.30
& 76.31
& 16.57
& 50.77
& 23.48
& 74.38
& 63.83
& 14.84
& 76.34
\\

& \textbf{\ours (Ours)}$^\dagger$
& 32.74
& \textbf{9.51}
& \textbf{16.35}
& \textbf{76.78}
& 18.17
& \textbf{56.76}
& \textbf{25.98}
& 78.73
& 70.53
& \textbf{17.96}
& \textbf{79.43}
\\

\bottomrule
\end{tabular}
}
\caption{
Full results on the FINDSum-Liquidity test set. Higher values are
better. $\dagger$ denotes methods using GPT-5-nano as the backbone.
Best results are shown in \textbf{bold}. All metrics are reported on a 0--100 scale.
}
\label{tab:app_main_results_full_liq}
\end{table*}

We begin with the analysis of overall performance. \ours achieves the strongest overall performance across
the two FINDSum subsets, although it does not outperform every individual metric. On FINDSum-ROO, it obtains the best result on 8/11 reported metrics, including all three ROUGE scores, BERTScore, NC, SummaC, G-Cov, and TGF. On FINDSum-Liquidity, it achieves the best R-2, R-L, BERTScore, NC, NS, G-Cov, and TGF. These results indicate that organizing source-grounded evidence and preserving cross-modal relations
improve analytical faithfulness without requiring a corresponding loss in reference-based summary quality.

\noindent \textbf{Numerical content selection.} Next, we analyze an important set of metrics (NP, NC, NS) as introduced in \citet{liu2022long}. We first compute numerical NP, NC, and NS for each instance in the test set; we then average each metric over the whole test set to get the reported scores. \ours achieves the highest NC on both subsets, reaching 54.17 on ROO and 56.76 on Liquidity. It recovers a larger proportion of the numerical content selected by the reference summaries. Its NP is not always the highest, suggesting that broader numerical coverage may include additional source-supported quantities that do not appear in the single reference summary. On ROO, this precision--coverage balance yields an NS of 27.03, which is nearly identical to the best score of 27.08 obtained by RAG-Sum. On Liquidity, \ours achieves the highest NS of 25.98.

\noindent \textbf{Effect of input modality.} We then shift our focus to analyze the impact of input modalities. The table-only methods generally obtain the lowest TGF and G-Cov scores. Although tables provide precise quantitative values, they contain less of the explanatory and contextual information needed to reconstruct the reference summaries. Text-only methods perform more strongly, reflecting the importance of narrative evidence. TextRank, in particular, remains competitive on several metrics, showing that extractive selection can preserve locally supported and reference-relevant information. Methods using both modalities generally achieve the strongest joint performance. This supports their complementary roles: tables provide exact quantities, whereas narrative passages describe their drivers, qualifications, and implications. The advantage of \ours over other joint-input methods further indicates that access to both modalities is not sufficient by itself; related evidence must also be correctly associated and organized before generation.

\noindent \textbf{Faithfulness--coverage trade-off.} We argue that TGF and G-Cov capture complementary properties. TGF averages faithfulness over the generated claims and therefore does not penalize information that a system never attempts to express. G-Cov instead averages grounded coverage over the reference claims, so omitted reference content receives little or no credit. A method with high TGF but low G-Cov is therefore generating a relatively small or conservative set of reliable claims rather than providing comprehensive grounded coverage. RAG-Sum illustrates this distinction. On ROO, it obtains a relatively high TGF of 69.47 but a G-Cov of only 13.10. The pattern is more pronounced on Liquidity, where it reaches a TGF of 76.34 but a G-Cov of 14.84. Its retrieval stage may expose the generator to a compact set of highly relevant evidence, improving the faithfulness of the claims that are produced. At the same time, restricting generation to the retrieved evidence can leave other reference claims uncovered. Direct-LC exhibits a different trade-off on Liquidity. Its G-Cov of 17.75 is close to the best score, indicating that it expresses a relatively broad range of reference content. However, its TGF of 69.29 is substantially below that of RAG-Sum and \ours. Providing the full source makes more information available, but it also increases the difficulty of correctly grounding quantities and associating them with the appropriate narrative interpretations. These comparisons demonstrate why TGF and G-Cov should be interpreted jointly. Moreover, \ours achieves the highest TGF and G-Cov on both FINDSum subsets, indicating that its faithfulness gains do not arise merely from generating fewer or more conservative claims.

\noindent\textbf{Analysis on the Summary Output Length.}
Table~\ref{tab:generated-summary-length} reports generated
summary lengths using the \texttt{o200k\_base} tokenizer.
We observe that \ours produces longer summaries on FINDSum-ROO,
averaging 561.6 tokens compared with 242.7--252.7 tokens
for the other joint text--table methods.
The difference is smaller on FINDSum-Liquidity, where
\ours averages 889.0 tokens versus 775.6--839.5 tokens.
In contrast, its USTT summaries are shorter, averaging
136.6 tokens compared with 154.2--175.9 tokens.

These results show that \ours does not uniformly
increase output length across datasets.
Nevertheless, the additional output space on FINDSum,
particularly ROO, may partly contribute to improved
coverage of quantitative facts and explanatory relations.
While longer outputs do not inherently increase
the claim-averaged TGF score, length-related differences
in content selection and claim composition should be
considered when interpreting the faithfulness results.

\begin{table*}[t]
\centering
\small
\setlength{\tabcolsep}{5pt}
\begin{tabular}{llrrr}
\toprule
Method & Input & FINDSum-ROO & FINDSum-Liquidity & USTT \\
\midrule
TextRank
& Text
& $280.4 \pm 33.7$ & $647.9 \pm 76.6$ & $112.0 \pm 36.4$ \\
BART
& Text
& $74.4 \pm 18.3$ & $76.5 \pm 20.5$ & $94.3 \pm 62.9$ \\
BigBird-Pegasus
& Text
& $118.5 \pm 76.0$ & $102.3 \pm 50.3$ & $59.9 \pm 34.2$ \\
\midrule
BART
& Table
& $54.1 \pm 14.4$ & $56.4 \pm 26.1$ & $109.5 \pm 89.3$ \\
BigBird-Pegasus
& Table
& $155.5 \pm 67.9$ & $151.5 \pm 62.0$ & $114.0 \pm 65.3$ \\
Table-RAG$^\dagger$
& Table
& $383.9 \pm 65.4$ & $309.4 \pm 47.9$ & $148.5 \pm 22.1$ \\
\midrule
GCG$^\dagger$
& Text + Table
& $252.7 \pm 18.8$ & $839.5 \pm 154.9$ & $172.5 \pm 23.5$ \\
Direct-LC$^\dagger$
& Text + Table
& $250.5 \pm 28.3$ & $822.1 \pm 100.2$ & $175.9 \pm 20.7$ \\
RAG-Sum$^\dagger$
& Text + Table
& $242.7 \pm 97.0$ & $775.6 \pm 166.7$ & $154.2 \pm 20.2$ \\
\textbf{\ours}$^\dagger$
& Text + Table
& $561.6 \pm 73.1$ & $889.0 \pm 131.1$ & $136.6 \pm 55.4$ \\
\bottomrule
\end{tabular}
\caption{
Generated summary length, reported as mean$\pm$std in BPE tokens using the \texttt{o200k\_base} tokenizer. Statistics are computed on the test set of each dataset. $^\dagger$ denotes methods using GPT-5-nano as the backbone.
}
\label{tab:generated-summary-length}
\end{table*}

\noindent \textbf{Comparison with generic factuality metrics.} Lastly, we provide analysis around generic factuality metrics we utilize in this work. AlignScore and SummaC produce rankings that do not always agree with TGF or G-Cov. On ROO, GCG obtains the highest AlignScore of 95.31, but its TGF and G-Cov are only 64.47 and 13.57, respectively. Text-only BART similarly obtains high AlignScore and SummaC scores of 94.63 and 81.82, while achieving substantially lower TGF and G-Cov scores of 54.33 and 10.73. The discrepancy is also visible on Liquidity, where text-only BART obtains the highest AlignScore and SummaC scores but reaches only 63.70 TGF and 10.56 G-Cov. Generic factuality metrics can therefore assign favorable scores to locally consistent summaries without reflecting whether a broad set of salient reference claims is covered or whether quantitative facts are associated with their correct narrative interpretations. TGF and G-Cov therefore provide complementary relation-sensitive faithfulness and grounded-coverage signals that are not fully captured by these generic scores.

\noindent \textbf{Results on USTT.} On USTT, we observe very similar patterns as in FINDSum. In Table~\ref{tab:main_results_full_ustt}, \ours achieves the best R-1, R-L, NP, NC, NS, AlignScore, and G-Cov results. Its advantage is particularly clear for numerical content selection, improving NC from 30.36 for the next-best method to 49.79 and NS from 21.44 to 27.36. TextRank obtains the highest TGF, likely benefiting from its conservative extractive behavior, but achieves a lower G-Cov. This contrast further demonstrates the complementarity of the two metrics: higher faithfulness among the claims expressed does not necessarily imply broader grounded coverage of the reference content. Among joint-input methods, \ours achieves the strongest results on both TGF and G-Cov.

\begin{table*}[t]
\centering
\scriptsize
\setlength{\tabcolsep}{2.2pt}
\renewcommand{\arraystretch}{0.96}

\resizebox{\textwidth}{!}{
\begin{tabular}{@{}llccccccccccc@{}}
\toprule
\textbf{Input}
& \textbf{Method}
& \textbf{R-1}
& \textbf{R-2}
& \textbf{R-L}
& \textbf{BS}
& \textbf{NP}
& \textbf{NC}
& \textbf{NS}
& \textbf{AlignScore}
& \textbf{SummaC}
& \textbf{G-Cov}
& \textbf{TGF} \\
\midrule

\multirow{3}{*}{\shortstack[l]{Text-\\only}}
& TextRank
& 24.77
& \textbf{10.20}
& 13.83
& 60.24
& 16.82
& 30.36
& 21.44
& 77.35
& 52.20
& 19.12
& \textbf{66.41}
\\

& BART
& 22.84
& 8.90
& 14.25
& 59.80
& 14.81
& 26.99
& 19.12
& 67.30
& 42.03
& 17.18
& 54.26
\\

& BigBird-Pegasus
& 24.18
& 7.69
& 14.17
& 60.12
& 16.19
& 27.51
& 20.35
& 64.42
& 47.83
& 16.74
& 56.91
\\
\midrule

\multirow{3}{*}{\shortstack[l]{Table-\\only}}
& BART
& 20.46
& 5.21
& 13.72
& 60.01
& 10.45
& 19.52
& 13.32
& 36.76
& 47.91
& 11.24
& 52.53
\\

& BigBird-Pegasus
& 20.71
& 5.54
& 12.01
& 59.25
& 12.33
& 18.89
& 13.98
& 46.16
& 49.80
& 11.45
& 55.79
\\

& Table-RAG$^\dagger$
& 19.26
& 5.66
& 10.77
& 61.50
& 13.31
& 12.74
& 12.18
& 73.28
& 42.44
& 10.16
& 56.13
\\
\midrule

\multirow{4}{*}{\shortstack[l]{Text +\\Table}}
& GCG$^\dagger$
& 24.83
& 8.93
& 12.92
& 63.84
& 19.50
& 24.21
& 18.76
& 64.77
& 42.31
& 16.19
& 59.71
\\

& Direct-LC$^\dagger$
& 25.14
& 9.37
& 13.26
& 63.71
& 17.76
& 25.78
& 18.63
& 74.55
& \textbf{54.70}
& 18.43
& 58.51
\\

& RAG-Sum$^\dagger$
& 24.56
& 9.91
& 12.93
& \textbf{64.47}
& 16.04
& 23.89
& 15.89
& 69.47
& 42.93
& 15.28
& 60.64
\\

& \textbf{\ours (Ours)}$^\dagger$
& \textbf{26.10}
& 9.18
& \textbf{15.16}
& 64.16
& \textbf{19.81}
& \textbf{49.79}
& \textbf{27.36}
& \textbf{77.99}
& 54.02
& \textbf{19.78}
& 63.71
\\

\bottomrule
\end{tabular}
}
\caption{
Full results on the USTT test set. Higher values are better.
$\dagger$ denotes methods using GPT-5-nano as the backbone.
Best results are shown in \textbf{bold}. All metrics are reported on a 0--100 scale.
}
\label{tab:main_results_full_ustt}
\end{table*}

\subsection{Cost and Latency Analysis} We analyze the inference cost of summary generation and automatic evaluation on FINDSum-ROO, FINDSum-Liquidity, and USTT. Costs are computed from API-reported token usage and the providers' published prices at the time of experimentation. 

\noindent\textbf{Summary Generation.} Using GPT-5-nano, \textsc{LoomSum} costs, on average, \$0.015, \$0.013, and \$0.008 per document on FINDSum-ROO, FINDSum-Liquidity, and USTT, respectively. The corresponding mean end-to-end latencies are 118.9, 105.2, and 89.4 seconds per document across three datasets. 

\noindent\textbf{TGF Evaluation.} TGF uses o3-mini with low reasoning effort for claim decomposition and scoring, with each claim evaluated three times using the five-point rubric. Its average cost is \$0.35 per summary on FINDSum-ROO, \$0.13 on FINDSum-Liquidity, and \$0.07 on USTT. Evaluation is therefore more expensive than generation, largely because it requires repeated claim-level judgments.

\section{Base-Model Sensitivity Analysis} \label{app:base_model_analysis}
\begin{figure*}[h]
\centering
\includegraphics[width=1.0\linewidth]{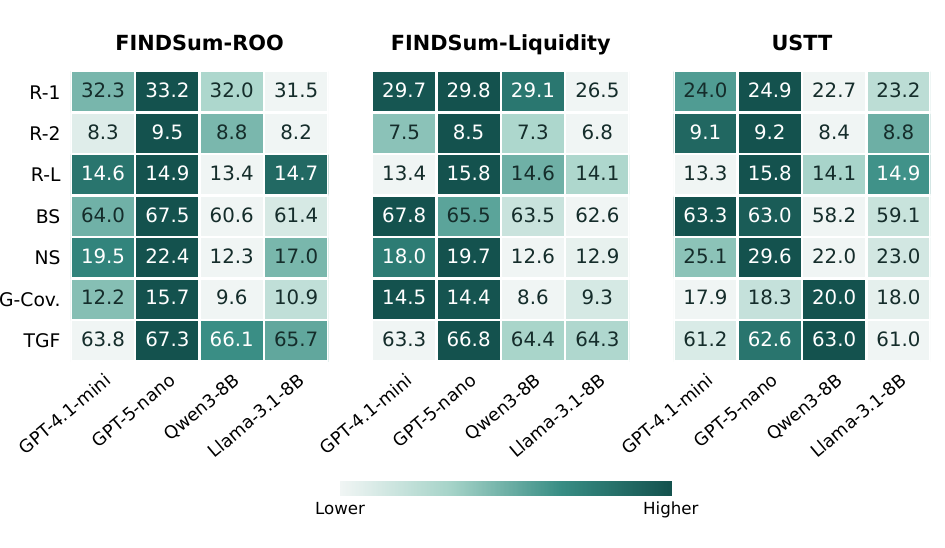}
\caption{Backbone sensitivity of \textsc{LoomSum} on
(Left.) FINDSum-ROO, (Middle.) FINDSum-Liquidity, and (Right.) USTT.
We report ROUGE-1/2/L (R-1/R-2/R-L), BERTScore (BS),
Number Selection (NS), grounded coverage (G-Cov.), and
Table-Grounded Faithfulness (TGF); higher is better.}
\label{app:fig_sen_analy}
\end{figure*}

To assess the sensitivity of \textsc{LoomSum} to its base LLM, we evaluate the same pipeline with four backbones on the same \textbf{500 randomly sampled} test samples from each dataset: GPT-4.1-mini, GPT-5-nano, Qwen3-8B~\cite{yang2025qwen3}, and Llama3.1-8B~\cite{grattafiori2024llama}. Figure~\ref{app:fig_sen_analy} reports R-1, R-2, R-L, BERTScore, NS, G-Cov., and TGF for FINDSum-ROO, FINDSum-Liquidity, and USTT. Among the proprietary models, GPT-5-nano achieves higher ROUGE, NS, and TGF scores than GPT-4.1-mini across all
three datasets, while their relative performance on BERTScore and G-Cov. varies by dataset. Among the two 8B open models, Qwen3-8B achieves higher TGF scores on all three datasets, whereas Llama3.1-8B achieves higher NS scores; neither consistently leads on ROUGE. TGF scores are more closely grouped across backbones
on USTT than on the two FINDSum subsets. Overall, these results demonstrate the applicability of \textsc{LoomSum} across the evaluated model families, and further show that \textsc{LoomSum} is relatively insensitive to the choice of base LLMs.

\section{More on Human Evaluation} \label{app:human_eval}

In this section, we provide more details on the human evaluation activity and the annotation procedures. Each annotator received the tasks in a different order. The annotation interface displayed an anonymized task identifier, the generated claim, and the retrieved table and narrative evidence. Method names, reference summaries, and automatic TGF scores were not shown. The instructions are provided below in the prompt box. Annotators first verify that the claim applies to the target dimension. NG evaluates whether quantitative content is supported with the correct entity, metric, value, unit, direction, and reporting period. AS evaluates whether an analytical statement, such as a trend, comparison, explanation, or qualification, is supported. RC evaluates whether a claim correctly associates a quantitative fact with its narrative interpretation. Annotators complete a binary checklist of the relevant criteria and then produce an overall rating from 1 to 5, following the same scale used by the automatic TGF evaluator. Next, we provide the results for Inter-annotator agreement. The agreement is computed from the original independent ratings, before any discussion. Cases of maximal disagreement, for example, one annotator assigning a rating of 1 and the other annotators assigning a rating of 5, are subsequently discussed to obtain a consensus rating for the correlation analysis.

\begin{table}[t]
\centering
\small
\setlength{\tabcolsep}{5.5pt}
\renewcommand{\arraystretch}{1.05}

\begin{tabular}{@{}lrrrr@{}}
\toprule
\textbf{Dimension}
& \textbf{$N$}
& \textbf{Exact}
& \textbf{Within $\pm 1$}
& \textbf{Ordinal $\alpha$} \\
\midrule
NG      & 167 & 69.5\% & 76.0\% & 0.595 \\
AS      & 171 & 71.9\% & 75.4\% & 0.479 \\
RC      & 139 & 74.8\% & 80.6\% & 0.532 \\
\bottomrule
\end{tabular}
\caption{
Exact agreement requires identical ratings from all three
annotators; within-1 agreement requires the maximum and minimum
ratings to differ by at most one point.
Each dimension is evaluated separately on applicable claims
with valid ratings from all three annotators.
Ordinal Krippendorff's $\alpha$ is computed from the same
independent, pre-adjudication ratings.
}
\label{tab:app_human_intra_agreement}
\end{table}

The results in Table~\ref{tab:app_human_intra_agreement}
show exact agreement ranging from 69.4\% to 74.8\%, and
within-one-point agreement ranging from 75.4\% to 80.6\%.
Ordinal Krippendorff's $\alpha$ ranges from 0.479 to 0.595,
with NG achieving the highest chance-corrected reliability,
followed by RC and AS. Although RC has the highest exact and
within-one-point agreement, its $\alpha$ is lower than that
of NG. These measures capture different aspects of agreement:
raw agreement describes how frequently ratings coincide or
differ by at most one point, whereas ordinal $\alpha$
additionally accounts for chance-expected disagreement and
the ordering of rating categories. Overall, the three-annotator evaluation exhibits broadly consistent ratings across dimensions, with stronger
chance-corrected agreement for NG and RC than for AS.  

\begin{promptbox}{Annotation Instructions}

Task.
For each example, evaluate whether the generated claim is supported by
the accompanying source evidence. The evidence packet may contain
table records, narrative passages, or both. Judge only against the
provided evidence. Do not use external knowledge, assume unstated
relations, or correct errors in the generated claim.

Step 1: Determine the claim type. Assign the claim to one of three categories.
- Numeric-only claims state a quantitative fact without attaching
an explanation or interpretation.
- Analysis-only claims state a trend, comparison, explanation,
risk, or interpretation without a central numerical assertion.
- Cross-source linked claims associate a quantitative fact with
an explanation, driver, offset, comparison, or interpretation.

Numeric-only claims are evaluated for Numeric Grounding (NG);
analysis-only claims are evaluated for Analysis Support (AS);
cross-source linked claims are evaluated for NG, AS, and Relation
Consistency (RC). Mark non-applicable dimensions as N/A.

Step 2: Evaluate Numeric Grounding.
For claims to which NG applies, determine whether the numerical value,
direction of change, metric, entity, reporting period, and unit are
correct. A number should not be marked correct merely because it appears
somewhere in the evidence. It must refer to the correct metric, entity,
unit, and reporting period. Numerically equivalent surface forms may be
treated as matching.

Step 3: Evaluate Analysis Support.
For claims to which AS applies, determine whether the evidence supports
the stated trend or comparison direction, the asserted attribution,
explanation, or mechanism, and the stated magnitude or strength of the
analysis. An interpretation that is plausible but not stated or licensed
by the evidence should be marked unsupported. Causal wording requires
stronger evidence than simple co-occurrence or correlation.

Step 4: Evaluate Relation Consistency.
For cross-source linked claims, evaluate whether the quantitative fact is
associated with the correct narrative interpretation. Determine whether
the number is paired with the correct explanation, whether the causal or
directional relation is preserved, whether both components concern the
same entity and reporting period, and whether the claim introduces an
unsupported inference.

NG, AS, and RC must be judged separately. A number and an explanation may
each be supported individually while their asserted relationship is
incorrect. Such a claim may receive positive NG and AS judgments but a
negative RC judgment.


Step 5: Overall Faithfulness.
Assign an independent overall faithfulness score.
A score of 5 indicates that the complete claim is fully
supported; 4 indicates substantial support with only a minor
omission or ambiguity; 3 indicates partial support with a
substantive ambiguity, omission, or local error; 2 indicates
that most of the claim is unsupported although some content is correct;
and 1 indicates that the central proposition is unsupported,
contradicted, or linked to the wrong evidence.

The overall score should reflect the faithfulness of the complete claim,
rather than an arithmetic average of the applicable dimensions.

\end{promptbox}

\section{More on case studies} \label{sec:app_case_study} 

We further examine two complementary case studies from FINDSum and USTT. The FINDSum example focuses on relation preservation when quantitative facts and their explanations are distributed across different parts of a report. The USTT example instead illustrates grounded omission, where methods generate source-supported content but fail to cover the claims emphasized by the reference summary.

Table~\ref{tab:case_study_FINDSum} presents an example from Commerce Bancshares' 2015 annual report. We show the source-supported relations, the reference summary, and verbatim excerpts from each method's output. Ellipses indicate omitted sentences. \textcolor{relationred}{\textbf{Red}} highlights only the phrase responsible for an unsupported relation, whereas {\color{ForestGreen}\textbf{Green}} highlights correctly preserved relations in the output.

\begin{table*}[t]
\centering
\footnotesize
\setlength{\tabcolsep}{4pt}
\renewcommand{\arraystretch}{1.12}

\begin{tabularx}{\textwidth}{
    @{}
    >{\raggedright\arraybackslash}p{0.11\textwidth}
    >{\raggedright\arraybackslash}X
    >{\raggedright\arraybackslash}p{0.24\textwidth}
    @{}
}
\toprule
\textbf{Source / Method}
&
\textbf{Evidence or output excerpt}
&
\textbf{Analysis}
\\
\midrule

\textbf{Source Document}
&
\textbf{R1:} Total revenue increased by \$25.7 million, comprising
\$11.6 million growth in non-interest income and \$14.1 million growth
in net interest income.
\textbf{R2:} Loan interest income increased by \$10.2 million because
average loans grew by \$609.0 million, partly offset by a
12-basis-point decline in rates.
\textbf{R3:} Net yield declined from 3.00\% to 2.94\% because of
continuing low interest rates.
&
The source establishes the revenue decomposition, the complete
loan-interest driver chain, and the cause of margin compression.
\\

\addlinespace

\textbf{Reference Summary}
&
Net income increased to \$263.7 million from \$261.8 million.
The increase resulted from \$14.1 million growth in net interest income
and \$11.6 million growth in non-interest income.
\ldots{}
These gains were partly offset by higher non-interest expense and lower
investment-security gains.
&
The reference preserves the main income decomposition, but omits the
loan-interest driver chain and the explanation for the yield decline.
\\

\midrule

Direct-LC
&
Total revenue increased \$25.7 million
\wronglink{because continuing low interest rates depressed net yield
to 2.94\%}.
\ldots{}
Net interest income was \$634.3 million, while non-interest income was
\$447.6 million, an increase of \$11.6 million.
\ldots{}
The output also reports provision, asset-quality, capital, dividend,
and shareholder-return statistics.
&
The revenue relation is explicitly stated but incorrectly linked to
the low-rate environment. Broad coverage introduces many secondary
indicators without preserving a focused driver-based narrative.
\\

\addlinespace

GCG
&
Total revenue increased \$25.7 million versus 2014, driven by
\$11.6 million growth in non-interest income and \$14.1 million growth
in net interest income.
\ldots{}
Net yield declined from 3.00\% to 2.94\%
\wronglink{because average loans grew by \$609.0 million, or 5.4\%}.
\ldots{}
Asset quality improved as net loan charge-offs and non-performing
assets declined.
&
R1 is explicitly preserved, but loan growth is incorrectly attached to
the yield decline. The positive and negative components of the
loan-interest driver chain are not coherently distinguished.
\\

\addlinespace

RAG-Sum
&
Total revenue increased \$25.7 million, reflecting \$11.6 million
growth in non-interest income and \$14.1 million growth in net interest
income.
\ldots{}
Net interest income increased \$14.1 million due to
\wronglink{higher trust fees, bank-card transaction fees, and
mortgage-banking revenue}.
\ldots{}
The output additionally discusses impaired securities, private-equity
investments, dividends, and shareholder return.
&
The revenue decomposition is preserved, but valid non-interest-income
drivers are assigned to net interest income. Retrieval broadens
coverage, yet also introduces secondary details and an incorrect
cross-category relation.
\\

\addlinespace

TextRank
&
Total revenue increased \$25.7 million due to growth in non-interest
income of \$11.6 million and growth in net interest income of
\$14.1 million.
\ldots{}
\wronglink{Loan interest income increased \$10.2 million}.
\ldots{}
Net interest income totaled \$664.0 million on a tax-equivalent basis,
an increase of \$15.4 million.
&
R1 is retained through sentence extraction, but the reason explained in R2 is not preserved in the output texts.
\\

\addlinespace

\textsc{LoomSum}
&
Total revenue increased \$25.7 million due to \$11.6 million growth in
non-interest income and \$14.1 million growth in net interest income.
\ldots{}
{\color{ForestGreen}Loan interest income increased \$10.2 million as average loans grew
by \$609.0 million; these improvements are partly offset by a 12-basis-point decline in rates and higher
non-interest expense.}
\ldots{}
Net yield declined from 3.00\% to 2.94\% because of continuing low
interest rates.
&
\ours explicitly preserves R1--R3, including the opposing effects of
loan growth and lower rates. It synthesizes the evidence into a
coherent narrative without introducing excessive secondary detail.
\\

\bottomrule
\end{tabularx}

\caption{
Case study of quantitative--narrative relation preservation. The source
document provides the grounding evidence, while the reference summary
is included for comparison. Each row shows selected method output excerpts.
\textcolor{relationred}{\textbf{Red}} highlights only the phrase
responsible for an unsupported or reversed relation. {\color{ForestGreen} \textbf{Green}} highlights the correct relations and interpretations of our method.
}
\label{tab:case_study_FINDSum}
\end{table*}

This example shows that covering the relevant quantities is not sufficient for analytical faithfulness. Direct-LC recovers several major indicators but incorrectly connects the revenue increase to the low-rate environment and introduces many secondary statistics. GCG and RAG-Sum preserve the revenue decomposition, yet attach valid drivers to the wrong outcomes: loan growth is associated with the yield decline, while non-interest-income drivers are assigned to net interest income. TextRank preserves substantial source wording but reverses the opposing effects of higher loan balances and lower rates. In contrast, \ours preserves the revenue decomposition, the complete loan-interest driver chain, and the explanation for the yield decline. In particular, it distinguishes the positive effect of loan growth from the offsetting effect of lower rates. The case therefore illustrates that the principal baseline failure is often not an unsupported individual fact, but an unsupported composition of otherwise valid facts and explanations.

\providecommand{\misszh}[1]{\textcolor{BrickRed}{#1}}
\providecommand{\keepzh}[1]{\textcolor{ForestGreen}{#1}}
\begin{table*}[t]
\begin{CJK*}{UTF8}{gbsn}
\centering
\setlength{\tabcolsep}{4pt}
\renewcommand{\arraystretch}{1.0}
\footnotesize
\begin{tabularx}{\textwidth}{
  @{}>{\raggedright\arraybackslash\bfseries}p{0.12\textwidth}
  >{\raggedright\arraybackslash}X
  >{\raggedright\arraybackslash}p{0.27\textwidth}@{}}
\toprule
Source\&Method &   Evidence or output excerpt & Analysis \\
\midrule
Source
& \textbf{R1：}美元兑人民币中间价上行405BP至6.7503。(Translated version: The USD/CNY central parity rate rose by 405 basis points (bps) to 6.7503.)
\textbf{R2：}在岸即期汇率上行619BP至6.7670。(Translated version: The onshore spot exchange rate rose by 619 bps to 6.7670.)
\textbf{R3：}在岸即期汇率较中间价高167BP。(Translated version: The onshore spot exchange rate was 167 bps higher than the central parity rate.)
\textbf{R4：}离岸即期汇率上行750.5BP至6.75875，较在岸低82.5BP。
配套表格则记录M2、M1和M0货币供应量。(Translated version: The offshore spot exchange rate rose by 750.5 bps to 6.75875, which was 82.5 bps lower than the onshore spot rate.)
& The reference-relevant FX evidence occurs in one source section; the paired
table is a cross-modal distractor. \\
\addlinespace
Reference
& 截至7月15日，美元兑人民币中间价较7月8日上行405BP至6.7503；
美元兑人民币即期汇率上行619BP至6.7670，较中间价高167BP；
美元兑离岸人民币即期汇率上行750.5BP至6.75875，较在岸即期汇率低82.5BP。(Translated version: As of 15 July, the USD/CNY central parity rate had risen by 405 basis points (bps) from 8 July to 6.7503; the USD/CNY onshore spot exchange rate had risen by 619 bps to 6.7670, 167 bps above the central parity rate; the USD/CNH offshore spot exchange rate had risen by 750.5 bps to 6.75875, 82.5 bps below the onshore spot exchange rate.)
& Preserves all four target claims and both cross-rate comparisons. \\
\midrule
Direct-LC
& 央行在7月11日-15日实施逆回购150亿元到期150亿元，同时以1000亿元MLF
投放并有同额MLF到期，净投放为零；\ldots{}货币市场方面，7月15日银行间
同业拆借加权利率较7月8日下行1.28基点至1.33\%，质押式回购下行0.58基点
至1.31\%；\ldots{} (Translated version: From 11 to 15 July, the People’s Bank of China (PBOC) conducted RMB 15 billion in reverse-repurchase operations, with RMB 15 billion maturing. It also injected RMB 100 billion through the medium-term lending facility (MLF), while an equal amount of MLF funds matured, resulting in zero net liquidity injection. In the money market, as of 15 July, the weighted average interbank lending rate had fallen by 1.28 basis points from 8 July to 1.33\%, while the pledged repo rate had declined by 0.58 basis points to 1.31\%; \ldots{})
& \misszh{The facts are source-supported but come from non-target sections;
all four reference-relevant FX claims are omitted.} \\
\addlinespace
GCG
& 央行在7月11日-15日实现净投放为零，逆回购投放150亿元、到期150亿元，
MLF投放1000亿元、到期1000亿元；\ldots{}货币市场方面，7月15日银行间
同业拆借加权利率较7月8日下降1.28基点至1.33\%，质押式回购利率下降
0.58基点至1.31\%；\ldots{} (Translated version: From 11 to 15 July, the People’s Bank of China (PBOC) recorded zero net liquidity injection. It conducted RMB 15 billion in reverse-repurchase operations, with RMB 15 billion maturing, and provided RMB 100 billion through the medium-term lending facility (MLF), with an equal RMB 100 billion of MLF funds maturing. In the money market, on 15 July, the weighted average interbank lending rate fell by 1.28 basis points (bps) from 8 July to 1.33\%, while the pledged repo rate declined by 0.58 bps to 1.31\%. \ldots{})
& \misszh{The output is grounded in the source, but verbalization does not repair
content selection; all target FX relations are omitted.} \\
\addlinespace
RAG-Sum
& 外汇方面，美元对人民币中间价和即期汇率均走高，提示短期汇率承压。(Translated version: In the foreign-exchange market, both the USD/CNY central parity rate and the spot exchange rate rose, indicating short-term pressure on the renminbi.)
& \misszh{Retrieves the FX topic but omits every target value and both
pairwise relations.} \\
\addlinespace
TextRank
& 货币市场：截至7月15日，银行间同业拆借加权平均利率较7月8日变动
-1.28BP至1.33\%，质押式回购加权平均利率较7月8日变动-0.58BP至
1.31\%；\ldots{} (Translated version: Money Market: As of 15 July, the weighted average interbank lending rate had decreased by 1.28 basis points (bps) from 8 July to 1.33\%, while the weighted average pledged repo rate had decreased by 0.58 bps to 1.31\%; \ldots{})
& \misszh{Extracts a source-faithful sentence from the wrong section; all four
reference-relevant FX claims are omitted.} \\
\addlinespace
\textbf{\ours}
& \keepzh{截止至7月15日，央行调整美元兑人民币中间价上行405BP至6.7503；在岸即期汇率上行619BP至6.7670，
较中间汇率价高167BP；离岸即期汇率上行750.5BP至6.75875，较在岸低82.5BP，呈跨市场小幅价差。(Translated version: As of 15 July, the People’s Bank of China (PBOC) set the USD/CNY central parity rate 405 basis points (bps) higher at 6.7503. The onshore spot exchange rate rose by 619 bps to 6.7670, standing 167 bps above the central parity rate. The offshore spot exchange rate increased by 750.5 bps to 6.75875, which was 82.5 bps below the onshore spot rate, indicating a small cross-market spread.)}
& \keepzh{\textbf{Preserves every value, direction, and relation.}} \\
\bottomrule
\end{tabularx}
\end{CJK*}
\caption{Case study on USTT. \ours preserves all four reference-relevant
claims; all baselines omit the target quantitative relations. Chinese excerpts
are copied from the original source and stored outputs.}
\label{tab:case_study_ustt}
\end{table*}

\noindent \textbf{USTT case. } The USTT example in Table~\ref{tab:case_study_ustt} illustrates a different failure mode: an output may remain source-grounded while providing poor coverage of the reference-relevant content. Direct-LC, GCG, and TextRank primarily select monetary-policy or money-market information from non-target sections. These statements are supported by the source, but they omit the four FX claims emphasized by the reference. RAG-Sum retrieves the correct topic and captures the coarse upward trend, yet removes the exact values and both cross-rate comparisons. \ours, by contrast, preserves the directions, values, and pairwise relations of all four target claims despite the presence of source-supported distractors. This case clarifies the role of G-Cov: source faithfulness alone does not guarantee that a summary covers the salient claims represented in the reference. Taken together, the FINDSum case demonstrates incorrect relation composition after partial content recovery, whereas the USTT case demonstrates grounded but incomplete content selection. \ours addresses both errors by explicitly aligning related evidence and organizing salient claims before generation.

\section{List of Prompts}
We here provide the complete prompt used to decompose generated summaries into atomic claims for TGF evaluation below. Code will be released after review.

\begin{promptbox}{TGF Claim Decomposition Prompt}
You are a claim-decomposition annotator for factuality evaluation. You will be provided with a summary in either English or Chinese. Your task is to decompose the summary exhaustively and extract it into minimal, independently verifiable factual claims for Table-Grounded Faithfulness evaluation.

Instructions:
1. First, read the summary carefully. Second, decompose the summary into factual claims.
2. You have to follow the rules below.

Rules:
1. Use only information explicitly stated in the generated summary. Do not introduce or infer external information.
2. You may resolve local pronouns or references using information within the summary, but do not add unstated entities, causes, or relations.
3. Preserve the exact semantic content of each claim, including:
   - values, units, entities, metrics, and reporting periods;
   - directions and magnitudes of change;
   - negation, attribution, modality, uncertainty, and hedging;
   - the stated strength of causal or explanatory relations.
4. Keep a quantitative proposition and an analytical proposition together only when the summary explicitly asserts a relation between them, such as a cause, driver, consequence, qualification, attribution, offset, or contextualization.
5. A linked claim may span multiple sentences when an explicit discourse cue or coreference establishes the relation, such as "this increase," "this decline," or "as a result." Pay more attention to this type of claim.
6. Extract all factual claims. Do not omit claims because they appear unimportant, repetitive, or difficult to verify.

Assign exactly one claim_type:
- numeric_only:
  An explicit quantitative assertion with no attached narrative interpretation. Quantitative comparisons that require multiple values may remain together as one claim. Dates or reporting periods alone do not make a claim numeric.

- analysis_only:
  A non-numeric trend, explanation, qualification, attribution, uncertainty statement, or interpretation that is independently asserted.

- cross_source_analytic:
  An explicit relation between a quantitative proposition and an analytical proposition. This label describes the structure of the generated claim and does not itself verify the provenance of either component.

For cross_source_analytic claims, separately record the quantitative component, analytical component, and relation type. For other claim types, return null for these fields.


Return only valid JSON using this schema:

{
  "claims": [
    {
      "claim_id": "c1",
      "claim": "minimal independently verifiable claim",
      "claim_type":
        "numeric_only|analysis_only|cross_source_analytic",
      "source_sentence_indices": [0],
      "source_span": "exact supporting span from the summary",
      "numeric_component": null,
      "analysis_component": null,
      "relation_type":
        null
    }
  ],
  "coverage_complete": true,
  "unprocessed_sentence_indices": []
}

Generated summary:
{numbered_summary}

\end{promptbox}

\begin{promptbox}{TGF LLM Judge Prompt}
You are a strict evaluator of Table-Grounded Faithfulness (TGF) for one generated claim, which presents in either English or Chinese. Use only the provided evidence packet. Do not use a reference summary, external knowledge, or plausibility. Do not correct or reinterpret the claim.

Claim-type hint:
{claim.claim_type}

Generated claim:
{claim.clean_claim_text}

Evidence packet:
{json.dumps(evidence_packet, ensure_ascii=False)}

Evaluate each applicable dimension independently:
- numeric_grounding:
  Whether the quantitative content is supported with the correct value, metric, entity, unit, direction, and reporting period.
- analysis_support:
  Whether the stated trend, comparison, explanation, qualification, attribution, or interpretation is supported.
- relation_consistency:
  Applies only when the claim links quantitative content with an analytical statement. Evaluate whether that specific association is supported. Individually supported components do not imply a
  supported relation.

Applicability:
- numeric_grounding applies to claims containing quantitative content.
- analysis_support applies to claims containing analytical content.
- relation_consistency applies only to explicit quantitative--analytical relations.

Rating rubric:
- 1: unsupported, contradicted, or substantially mismatched.
- 2: mostly unsupported; only weakly or indirectly related evidence.
- 3: partially or ambiguously supported.
- 4: mostly supported, with minor ambiguity or omitted qualification.
- 5: fully supported by the evidence.

If the evidence packet is insufficient to judge an applicable dimension, set "evidence_sufficient" to false and "rating" to null. For a non-applicable dimension, set "applicable" to false and "rating" to null.

Return only valid JSON:

{{
  "numeric_grounding": {{
    "applicable": true,
    "evidence_sufficient": true,
    "rating": 1,
    "rationale": "Brief evidence-based justification."
  }},
  "analysis_support": {{
    "applicable": true,
    "evidence_sufficient": true,
    "rating": 1,
    "rationale": "Brief evidence-based justification."
  }},
  "relation_consistency": {{
    "applicable": true,
    "evidence_sufficient": true,
    "rating": 1,
    "rationale": "Brief evidence-based justification."
  }}
}}
\end{promptbox}

\section{Usage of AI Assistants}
ChatGPT was used to assist with language editing, improving clarity and concision, and providing feedback on manuscript organization and presentation. All feedback and suggestions were independently evaluated by the authors. The authors conducted the experiments, verified all technical claims and citations, and take full responsibility for the final manuscript.

\end{document}